%% file: main.tex
\documentclass[10pt,twocolumn,letterpaper]{article}

\usepackage{iccv}              % To produce the CAMERA-READY version
\usepackage{graphicx}
\usepackage{afterpage}
\usepackage{placeins}
\usepackage{multirow}
\usepackage{colortbl, xcolor}
\definecolor{tablebg}{rgb}{0.95, 0.95, 0.98} % 背景色
\definecolor{sota}{rgb}{1, 0.5, 0.5}
\definecolor{secondbest}{rgb}{1, 0.75, 0.5} 
\usepackage{float}
\usepackage{amsmath}
\usepackage{graphicx} % 用于使用 \scalebox
\newcommand{\ourmodel}{WildSeg3D}
\newcommand{\myEq}[1]{Eq. \ref{#1}}


\definecolor{iccvblue}{rgb}{0.21,0.49,0.74}
\usepackage[pagebackref,breaklinks,colorlinks,allcolors=iccvblue]{hyperref}

\def\paperID{3022} % *** Enter the Paper ID here
\def\confName{ICCV}
\def\confYear{2025}

\title{WildSeg3D: Segment Any 3D Objects in the Wild from 2D Images}

\author{
    Yansong Guo\textsuperscript{1} \\
    \and
    Jie Hu\textsuperscript{2} \\
    \and
    Yansong Qu\textsuperscript{1} \\
    \and
    Liujuan Cao\textsuperscript{\dag}\textsuperscript{1} \\
    \and
    \textsuperscript{1}Xiamen University, 
    \textsuperscript{2}National University of Singapore
}

\begin{document}
\maketitle
\input{sec/0_abstract}
\input{sec/1_intro}
\input{sec/2_relatedwork}

\input{sec/3_methods}
\input{sec/4_experiments}

\input{sec/5_conclusion}

{
    \small
    \bibliographystyle{ieeenat_fullname}
    \bibliography{main}
}
\input{sec/X_suppl}

\end{document}

%% file: sec/0_abstract.tex
\begin{abstract}
Recent advances in interactive 3D segmentation from 2D images have demonstrated impressive performance.
However, current models typically require extensive scene-specific training to accurately reconstruct and segment objects, which limits their applicability in real-time scenarios.
In this paper, we introduce WildSeg3D, an efficient approach that enables the segmentation of arbitrary 3D objects across diverse environments using a feed-forward mechanism.
A key challenge of this feed-forward approach lies in the accumulation of 3D alignment errors across multiple 2D views, which can lead to inaccurate 3D segmentation results.
To address this issue, we propose Dynamic Global Aligning (DGA), a technique that improves the accuracy of global multi-view alignment by focusing on difficult-to-match 3D points across images, using a dynamic adjustment function.
Additionally, for real-time interactive segmentation, we introduce  Multi-view Group Mapping (MGM), a method that utilizes an object mask cache to integrate multi-view segmentations and respond rapidly to user prompts.
WildSeg3D demonstrates robust generalization across arbitrary scenes, thereby eliminating the need for scene-specific training.
Specifically, WildSeg3D not only attains the accuracy of state-of-the-art (SOTA) methods but also achieves a $40\times$ speedup compared to existing SOTA models.
Our code will be publicly available.
\end{abstract}

%% file: sec/1_intro.tex
\section{Introduction}
\label{sec:intro}

Interactive 3D segmentation from 2D images plays a critical role in 3D scene understanding and remains a fundamental challenge in computer vision, attracting significant attention from the research community~\cite{ISRF,kobayashi2022decomposing,SGISRF,tschernezki2022neural,peng2023openscene,takmaz2023openmask3d,LERF}.
This technology has broad applications across various fields, including virtual and augmented reality, real-time interactive systems, and automatic labeling.
Recent advancements in interactive 3D segmentation have demonstrated remarkable performance, particularly based on Neural Radiance Fields (NeRF)~\cite{nerf} and 3D Gaussian Splatting (3DGS)~\cite{3dgs}.               
For instance, models such as SA3D~\cite{sa3d} and SANeRF-HQ~\cite{liu2024sanerf} integrate NeRF with foundational segmentation models like Segment Anything Model (SAM)~\cite{sam}, aligning semantic information with 3D representations to enable effective 3D object segmentation.
Similarly, 3DGS-based approaches~\cite{semanticAnyIn3dgs,gaussianGroup,feature3dgs,qin2024langsplat,SAGA} address the high training demands of NeRF by constructing Gaussian feature fields in combination with SAM, facilitating faster model training.
However, both NeRF-based and 3DGS-based methods typically rely on extensive scene-specific training to obtain accurate 3D priors, which significantly hinders their applicability in real-time scenarios.
% 这里不用介绍下面这些工作
% These challenges underscore the need for 3D segmentation methods that avoid scene-specific training while distinguishing target features in real-time. Unlike NeRF and 3DGS, DUSt3R~\cite{wang2024dust3r} introduces a novel paradigm, supporting dense and unconstrained stereo 3D reconstruction from multi-view images without training on specific scenes, thus unifying various 3D vision tasks. MASt3R~\cite{mast3r}, an enhancement of DUSt3R, formulates image matching as a 3D task and introduces a novel head architecture along with a fast reciprocal matching scheme, significantly improving both image matching and 3D reconstruction efficiency. This paradigm shift opens pathways for integrating additional post-processing techniques, enabling real-time interactive 3D segmentation in diverse, unstructured environments.

%%%%%%%%%%%%%%% Figure 1.
\begin{figure*}[t]
  \centering
  \includegraphics[width=0.95\linewidth]{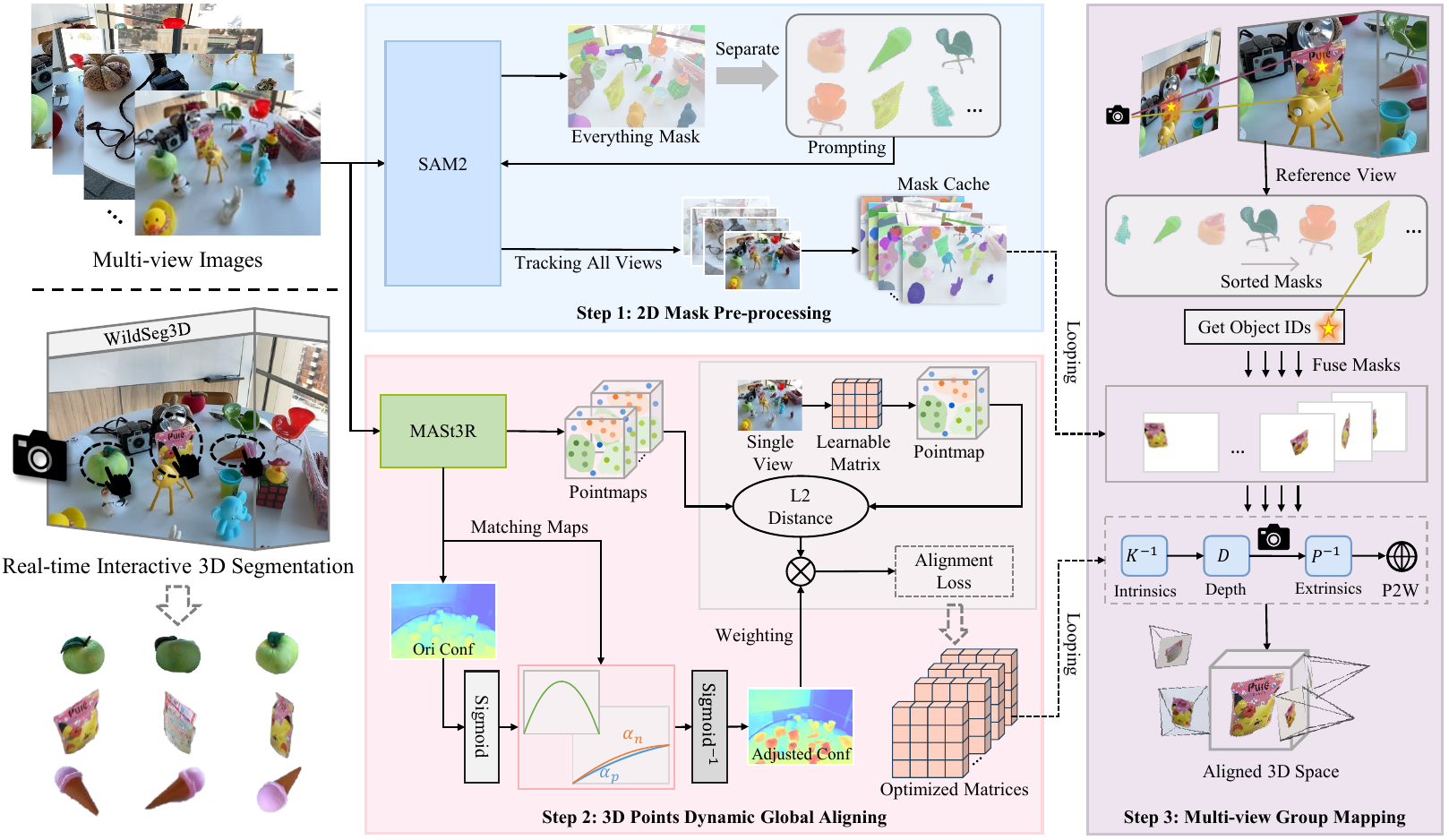}
  \vspace{-2.5mm}
  \caption{\textbf{Framework of~\ourmodel.} WildSeg3D operates in three stages.
  First, during the pre-processing stage, 2D feature masks are constructed from multi-view images using SAM2, providing offline support for interactive segmentation.
  Second, in the Dynamic Global Alignment (DGA) stage, a dynamic weight adjustment strategy is applied to achieve global alignment of the pointmaps generated by MASt3r, thereby improving the accuracy of object reconstruction.
  Finally, in the Multi-view Group Mapping (MGM) stage, multi-view masks for the target are retrieved from the mask cache based on user input, and these masks are transformed into an aligned 3D space in real time, where "P2W" refers to the transformation from pixel coordinates to the aligned world coordinates.}
\label{fig:model}
\end{figure*}

%%%
To overcome this limitation, we propose WildSeg3D, a novel approach with a feed-forward manner, eliminating the need for scene-specific training.
Inspired by Dust3r~\cite{wang2024dust3r} and Mast3r~\cite{mast3r}, our feed-forward approach represents 3D scenes as pointmaps and performs scene reconstruction via global alignment.
A key challenge in this process is the influence of redundant background points and the difficulty of matching 3D points across different views, which accumulates 3D alignment errors across multiple 2D views.
This can lead to inaccurate 3D segmentation with confusion between background and target objects. 
% However, a major challenge in this process arises from the influence of redundant background points and the difficulty of matching 3D points from different views, leading to blurred details and potential background confusion when matching the same target points across different views. %[maybe you should show some examples in ablation studies to support this claim?]
% Unlike 3D representations based on NeRF and 3DGS, we resort to MASt3R, a feed-forward approach that represents 3D scenes as pointmaps and performs scene reconstruction through global alignment.
% As a feed-forward method, directly aligning MASt3R's point clouds into a unified coordinate system introduces "global errors," leading to blurred details and potential background confusion when matching the same target points across different views.
%
%A major challenge in this process arises from the influence of redundant background points and the difficulty of matching 3D points from different views. 
To address this, we introduce Dynamic Global Alignment (DGA), a method that dynamically adjusts attention on view-specific points during matching, minimizing alignment errors during the global registration of pointmaps across multiple views.
For real-time interactive segmentation, we also propose a mask cache constructed during preprocessing, leveraging the multi-frame segmentation capabilities of SAM2~\cite{sam2}.
This cache stores consistent object masks across all viewpoints, providing offline data that can be used for subsequent real-time segmentation.
We further introduce Multi-view Group Mapping (MGM), which combines the mask cache with the DGA strategy to integrate multi-view segmentation results into an aligned global coordinate system.
The MGM module retrieves target feature points from the mask cache based on user inputs and generates a unified 3D mask for the target across all viewpoints.

WildSeg3D demonstrates robust generalization across diverse scenes, transforming 2D target masks into an aligned 3D space, thus enabling accurate 3D segmentation without the need for scene-specific training.
We conduct extensive experiments on WildSeg3D using complex real-world scenes to evaluate both segmentation accuracy and efficiency.
Our results show that WildSeg3D not only matches the accuracy of state-of-the-art (SOTA) methods~\cite{nvos,spinnerf,ISRF,SGISRF,sa3d,SAGA,ying2024omniseg3d,shen2024flashsplat} but also achieves a 40$\times$ speedup compared to the SOTA models.
Specifically, WildSeg3D completes scene reconstruction in under 30 seconds, a significant reduction compared to the fastest model, SA3D, which requires 780 seconds.
Additionally, real-time interaction results are delivered in just 5-20 milliseconds.

Our main contributions can be summarized as follows:
\begin{itemize}
    \item We introduce WildSeg3D, the first feed-forward 3D segmentation model that operates directly from 2D views, eliminating the need for scene-specific training and enabling immediate segmentation of arbitrary 3D objects in diverse environments.
    
    \item we propose Dynamic Global Alignment (DGA), a novel method for addressing 3D alignment errors and enhancing segmentation accuracy. We also propose Multi-view Group Mapping (MGM) that enables real-time interactive 3D segmentation with robust generalization across diverse scenes.
    
    \item Extensive experiments on complex real-world scenes demonstrate that WildSeg3D achieves both high segmentation accuracy and significant efficiency, processing 2D views to 3D segmentation over 40$\times$ faster than exisiting SOTA models.
\end{itemize}

%Contributions:
% 1. 我们介绍了WildSeg3D, 首个feed-forward的3D分割模型from 2D views。WildSeg3D避免了scene-specific training，实现了immediately segment any 3D objects in the wild from 2D views。

%% file: sec/2_relatedwork.tex
\section{Related Work}
\label{sec:relatedwork}
\subsection{2D Image-based 3D Reconstruction}
Recent advancements in image-based 3D reconstruction using neural networks have led to significant progress.
%NeRF-based
Innovations like Neural Radiance Fields (NeRF)~\cite{nerf} have shown strong performance in generating realistic novel viewpoints for view synthesis.
However, the reliance on neural networks in NeRF-based methods~\cite{wang2023rip,qu2023sg,sa3d,liu2024sanerf,wang2025rise,wang2024scarf} leads to long training and rendering time.
To improve surface reconstruction,~\cite{fu2022geo,guo2023streetsurf,long2022sparseneus,wang2021neus,wang2022hf} leverage the signed distance function (SDF) for surface representation and introduce an innovative volume rendering technique to learn an SDF model.
% 3DGS-based
Kerbl \textit{et al.} introduced 3D Gaussian Splatting (3DGS)~\cite{3dgs}, which provides an explicit representation of 3D scene information, bypassing the time-consuming implicit reconstruction process of NeRF via MLP.
Some studies~\cite{LERF,feature3dgs,gaussianGroup,qin2024langsplat,qu2024goi}, incorporate semantic information into 3DGS, equipping it with semantic awareness through training.
Other researches~\cite{luiten2023dynamic,wang2024gflow,yang2023real,yang2024deformable,qu2025drag} extend 3DGS by incorporating deformation fields to track the positions of 3D Gaussians at each timestamp, capturing dynamic 3D environments.
% Additional studies~\cite{chen2024text,ling2024align,ren2023dreamgaussian4d,tang2023dreamgaussian,yi2023gaussiandreamer,yin20234dgen} combine 3DGS with diffusion models to generate 3D objects.
%
% Multi-view stereo reconstruction (MVS) in the wild requires the estimation of camera parameters, including both intrinsic and extrinsic parameters, as a first step.
% DUSt3R - MASt3R
DUSt3R~\cite{wang2024dust3r} presents a novel approach for dense, unconstrained 3D reconstruction from arbitrary image collections without requiring camera calibration or viewpoint poses, unlike 3DGS and NeRF, which depend on dense viewpoints for scene construction. 
Building on DUSt3R, MASt3R~\cite{mast3r} reframes image matching as a 3D task, achieving superior 3D reconstruction performance. 
We extended MASt3R’s feed-forward mechanism to enable 3D scene perception and address alignment errors.

\subsection{2D Foundation Models}
Foundation models (FMs) have emerged as a transformative paradigm in AI. These models are typically trained on extensive datasets, possess a large number of parameters, and demonstrate adaptability across a broad spectrum of downstream tasks. Specifically, 2D visual foundation models (VFMs)~\cite{sam,sam2,wang2023seggpt,oquab2023dinov2,radford2021learning} have gained significant attention due to their ability to process and understand visual data.
Kirillov \textit{et al.} proposed the Segment Anything Model (SAM)~\cite{sam}, a 2D segmentation foundation model for prompt-based segmentation. SAM generates segmentation masks based on prompts that identify target objects in an image, allowing it to generalize across unseen categories.
As the successor to SAM, Segment Anything 2 (SAM2)~\cite{sam2} unifies video and image segmentation by utilizing a larger training dataset and incorporating architectural enhancements to improve performance across a wide range of tasks.
Our method leverages SAM2 to segment 2D images, maintaining consistent masks across multiple viewpoints, thereby addressing the challenges of real-time 3D interactive segmentation.

\subsection{Interactive 3D Segmentation}
Few approaches \cite{ISRF, kontogianni2023interactive, yue2023agile3d, lang2024iseg} support interactive segmentation directly in 3D space. For example, \cite{kontogianni2023interactive} directly segment 3D point clouds based on user clicks.
Meanwhile, representations in 2D image-based 3D reconstruction have driven progress in interactive 3D segmentation from 2D images. Inspired by the advancements in 3D Neural Scene Representation~\cite{nerf,3dgs}, several studies \cite{chen2023gnesf, inplace, liu2023instance, contrastiveLift, panopticLift, giraffe, yu2021unsupervised, sa3d} have explored 3D segmentation within these frameworks. 
% \cite{chen2023gnesf, inplace} attempt to encode 2D semantic masks into the neural radiance field, using 2D semantic supervision to achieve 3D segmentation during training.
% Additionally, several supervised \cite{liu2023instance, contrastiveLift, panopticLift} and unsupervised methods \cite{giraffe, yu2021unsupervised} have been proposed for 3D instance segmentation with NeRF. 
With SAM \cite{sam}, SA3D \cite{sa3d} proposed an automatic strategy for cross-view prompt collection, leveraging SAM to obtain 2D masks and guide 3D feature training. 
Within the context of 3DGS,  Feature3DGS \cite{feature3dgs} converts features from SAM's encoder into 3D space and uses SAM's decoder to generate masks.
Mask-lifting-based methods directly map 2D segmentation masks from SAM into 3D space. Notable examples include SAGA \cite{SAGA}, Gaussian Grouping \cite{gaussianGroup}, SAGS \cite{semanticAnyIn3dgs}, Click-Gaussian \cite{choi2024click}, and FlashSplat \cite{shen2024flashsplat}.
These approaches computational costs of feature-to-mask conversion but still suffer from slow reconstructing. Our method distinguishes itself by achieving feed-forward segmentation, without scene-specific training.

%% file: sec/3_methods.tex
\section{Methods}
\subsection{Preliminary: Feed-Forward Mechanism}
The feed-forward mechanism for 3D scene reconstruction was initially proposed in DUSt3R~\cite{wang2024dust3r} and further refined in MASt3R~\cite{mast3r}.
Unlike methods such as NeRF~\cite{nerf} and 3DGS~\cite{3dgs}, which require scene-specific pre-training, the feed-forward approach enables general 3D scene reconstruction through two key steps: pointmap prediction and global alignment.

\noindent\textbf{Pointmap Prediction.}
The process of pointmap prediction can be described as a network function $\mathcal{F} :(I^n,I^m) \to (X^{n,e},C^{n,e},F^{n,e},X^{m,e},C^{m,e},F^{m,e})$, the inputs are two RGB images $I^n,I^m\in \mathbb{R}^{W\times H\times 3}$ from different views of the scene, the outputs include two corresponding pointmaps, $X^{n,e},X^{m,e}\in \mathbb{R}^{W\times H\times 3}$, confidence maps $C^{n,e},C^{m,e}\in \mathbb{R}^{W\times H}$, and dense local features $F^{n,e},F^{m,e} \in \mathbb{R}^{W \times H \times d}$. Note that $e=(n, m)$ refers to the image pair formed by $I^n$ and $I^m$, and both pointmaps are positioned in the camera coordinate system of $I^n$.
The predicted pointmaps locate the 3D positions for every pixel of the input 2D images.

\noindent\textbf{Global Alignment.}
Global alignment is used as a post-process that optimizes the pointmaps from multiple views into an aligned 3D coordinate system.
Given a set of images $\left \{I^1,I^2,...,I^N \right \} $ from a scene, a connectivity graph $\mathcal{G} =(\mathcal{V},\mathcal{E})$ is constructed, where the vertices $\mathcal{V}$ represent the $N$ images, and each edge $e=(n,m)\in \mathcal{E}$ connects an image pair $I^n$ and $I^m$.
By traversing the connected graph $\mathcal{G}$, globally aligned pointmaps $\left \{\chi^n \in \mathbb{R}^{W\times H\times 3} \right \}$ are recovered for all pixel coordinates $(i,j) \in \left \{1...W \right \} \times \left \{1...H \right \}$ and all cameras for different views $n = 1, \ldots, N$.
The global optimization process is formulated as follows:
\begin{equation}
\chi^* = \arg\min_{\chi, P, \sigma} \sum_{e \in \mathcal{E}} \sum_{v \in e} \sum_{i=1}^{HW} C_i^{v,e} \left\| \chi_i^{v} - \sigma_e P_e X_i^{v,e} \right\|,
\label{method:globalAlignment}
\end{equation}
where $P_e \in \mathbb{R}^{3\times4}$ represents the pairwise pose, which is a rigid transformation used to align the pointmaps $X^{n,e},X^{m,e}$ with the world-coordinate pointmaps $\chi^n, \chi^m$.
Additionally, $\sigma_e$ is a scale factor, subject to the constraint that $ {\textstyle \prod_{e}\sigma_e=1}$ for all $e \in \mathcal{E}$.

\subsection{Task Formulation: Segment with Pointmaps}
%       郭岩松    已修改%%%%%%%%%%%%%%%%%%%
Although the feed-forward approach offers the advantage of generalizing across various scenes without the need for scene-specific pre-training, its application to 3D segmentation often results in lower accuracy.
This is primarily due to the accumulation of 3D alignment errors during the global alignment stage.
Given multiple 2D images from different viewpoints and a user-provided prompt specifying a target object in 2D, our task is to optimize the alignment of 3D object segmentation across these views. 
The goal is to transform the 2D object masks, related to the provided prompt, into an aligned 3D space.
Two key challenges need to be addressed in this task: (1) obtaining real-time segmentation masks of the target across all viewpoints based on user prompts, and (2) refining the alignment loss function to enhance segmentation accuracy, which requires a strategy to minimize discrepancies between predicted pointmaps from multiple views and globally aligned pointmaps.

\subsection{WildSeg3D}
As illustrated in~\Cref{fig:model}, WildSeg3D is a feed-forward framework designed for real-time interactive 3D segmentation from 2D views.
The framework operates in three stages: 2D mask pre-processing, 3D point dynamic global alignment, and multi-view group mapping.
In the 2D mask pre-processing stage, segmentation masks are generated by SAM2 from the input multi-view images and stored in a mask cache for efficient access during real-time segmentation.
In the 3D point dynamic global aligning stage, the proposed DGA refines the alignment of 3D pointmaps by focusing on challenging pixel correspondences across different views.
This approach ensures more accurate 3D scene reconstruction by addressing misalignments caused by complex or occluded regions.
In the multi-view group mapping stage, the stored masks are retrieved from the mask cache and the transform matrix learned through DGA is applied to map the multi-view pointmaps into an aligned 3D coordinate system in real time.
This process allows for the accurate 3D segmentation results based on user prompts.
\subsubsection{2D Mask Pre-processing}
In the 2D mask pre-processing stage, we eliminate the need for online computation during 2D segmentation by introducing a mask cache, enabling rapid, accurate segmentation across multiple viewpoints.
First, by leveraging the video tracking capabilities of SAM2, we perform panoptic segmentation on a single viewpoint, generating precise masks for each target object within the scene.
These object masks are then stored offline in the mask cache, creating a repository of segmentation data that can be efficiently accessed during real-time processing.
In subsequent stages, these pre-generated masks serve as accurate prompts for SAM2's tracking functionality, allowing consistent tracking and segmentation of each object across multiple viewpoints in later frames.
This approach ensures continuity and coherence in the segmentation process, even as the viewpoint changes.
By pre-generating and storing the masks offline, the mask cache reduces the computational burden at runtime.
Furthermore, this offline caching mechanism not only accelerates the segmentation process but also enhances its robustness.
\begin{figure*}[t]
  \centering
  \includegraphics[width=0.95\linewidth]{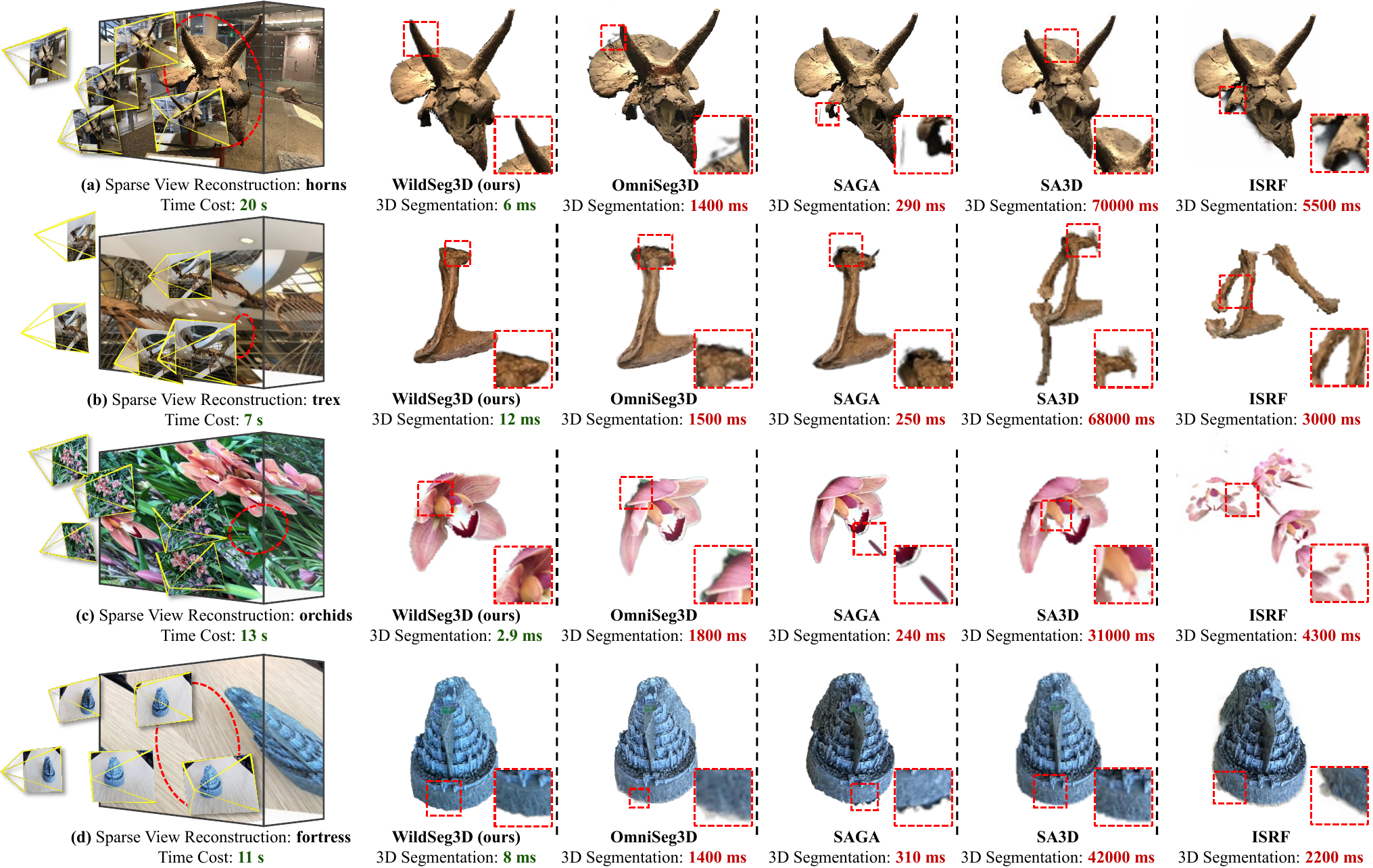}
  \vspace{-2.5mm}
  \caption{\textbf{Visualization on the NVOS dataset.} (a)-(d) show the sparse view reconstruction and  timing results on horns, trex, orchids, and fortress scenes, including preprocessing and DGA-based scene reconstruction. Target objects for segmentation are marked with red dashed lines in the first column. From left to right: segmentation results and elapsed time from prompt input to segmentation acquisition across models.}
\label{fig:qualitative1}
\end{figure*}

\subsubsection{3D Point Dynamic Global Aligning}
While the segmentation masks from the first step can be globally aligned to unify the 3D pointmaps into a single world coordinate system, challenges remain in achieving precise 3D segmentation across different viewpoints.
Specifically, misalignment and loss of detail can occur due to cluttered and inconsistent backgrounds in images from various viewpoints, weakening the alignment of target objects during the global alignment process.
To address these issues, we propose the Dynamic Global Aligning (DGA) method, which introduces soft masks to minimize background interference and dynamically adjusts aligning weights.
This adjustment enhances the focus on challenging sample points, thus improving the overall alignment effectiveness for the target objects.

\noindent\textbf{Soft-mask and Confidence Aggregation.}
As described in Eq.~\ref{method:globalAlignment}, the original global alignment approach considers all pixels from all viewpoints for alignment.
However, due to significant background differences between images from different viewpoints, the alignment process suffers from two issues: (1) background features are difficult to be aligned, and (2) the large proportion of the background in the images weakens the alignment focus on the objects to be segmented.
To mitigate this, we propose to soften the masks generated by SAM2 and store them in the mask cache.
Concretely, begin by multiplying the confidence of the pointmaps, $C\in \mathbb{R}^{W\times H}$, with the corresponding soft masks, $S\in \mathbb{R}^{W\times H}$, to obtain a weighted confidence.
A sigmoid function, $\sigma(\cdot)$, is then applied to the weighted confidence to map it to a confidence score in the range of 0 to 1.
For all views $v = 1, \ldots, N$, this process is expressed as follows:
\begin{equation}
F_i^{v, e} = \sigma(S_i^v\times C_i^{v, e}) = \frac{1}{1 + e^{-S_i^v\times C_i^{v, e}}},
\label{method:softmask}
\end{equation}
where $F\in \mathbb{R}^{W\times H}$ represents the adjusted confidence after applying the soft mask.

\noindent \textbf{Transform Matrix.}
%介绍K D P组成了Transform Matrix
We initialize a transform matrix for each view, composed of camera intrinsics, extrinsics, and depth information, enabling projection from pixel coordinates to world coordinates.
Given a 2D point $(x, y)$ from a single view, with depth map $\mathbf{D} \in \mathbb{R}^{W\times H}$, camera intrinsic matrix $\mathbf{K}\in\mathbb{R}^{3\times3}$, and extrinsic matrix $\mathbf{P}$, its corresponding 3D point $\mathbf{p}=(x,y,z)^T$ in world coordinates can be estimated as:
\begin{equation}
\mathbf{p} = \mathbf{P}^{-1} \mathbf{K}^{-1} 
\begin{pmatrix}
x \cdot \mathbf{D}(x,y) \\
y \cdot \mathbf{D}(x,y) \\
\mathbf{D}(x,y)
\end{pmatrix}.
\label{method:transform}
\end{equation}

For all views $v = 1, \ldots, N$, each 3D point $\chi^v_i$ in the world coordinate, where $i \in \mathbb{R}^{HW}$, can be derived using the above transform matrix.

\noindent \textbf{Dynamic Aligning Loss.} 
To handle challenging target sample points, we perform point pre-matching on all image pairs in the connectivity graph $\mathcal{G}$.
For each image pair, we generate a matching map $\Phi^{v, e} \in \mathbb{R}^{W\times H}$, which indicates whether each sample point $i \in \Phi^{v, e}$ has a matched point.
Based on the confidence scores derived from~\myEq{method:softmask}, we define a dynamic adjustment function as follows:
\begin{equation}
A_i^{v, e} = \frac{F_i^{v, e}  + \alpha_i^{v,e} \cdot F_i^{v, e}  \cdot (1 - F_i^{v, e} )}{1 + |\alpha_i^{v,e}| \cdot F_i^{v, e}  \cdot (1 - F_i^{v, e}) + \epsilon},
\label{method:dynamic}
\end{equation}
where $F_i^{v, e} \cdot (1 - F_i^{v, e} )$ enhances attention on difficult-to-match points, with confidence scores approaching 0.5, the axis of symmetry of the quadratic function.
The denominator ensures that the adjusted confidence score $A_i^{v, e}$ remains within an effective range.
The adjustment factor $\alpha$ is defined as:
\begin{equation}
\alpha_i^{v, e} = \begin{cases} 
\alpha_p & \text{if } \Phi_i^{v, e} = 1 \\
-\alpha_n & \text{if } \Phi_i^{v, e} = 0
\end{cases},
\label{method:alpha}
\end{equation}
where $\alpha_p$ is the positive adjustment factor for matching points, and $\alpha_n$ is the negative adjustment factor for non-matching points. Both are used to differentiate the weights between matching and non-matching points.

To map the dynamically adjusted confidence score back to its original value range, we apply the inverse of the sigmoid function, $\sigma^{-1}(\cdot)$, as follows:
\begin{equation}
W_i^{v, e} = \sigma^{-1}({A}_i^{v, e}) = -\ln(\frac{1}{{A}_i^{v, e}} - 1).
\label{method:sigmoidInverse}
\end{equation}

Finally, we compute the L2 distance between pointmaps of all image pairs and their corresponding pointmaps $\chi^v$ in world coordinates. The final Dynamic Global Aligning is defined as: 
\begin{equation}
\chi^* = \arg\min_{\chi, P, \sigma} \sum_{e \in \mathcal{E}} \sum_{v \in e} \sum_{i=1}^{HW} W_i^{v,e} \left\| \chi_i^{v} - \sigma_e P_e X_i^{v,e} \right\|,
\label{method:finalLoss}
\end{equation}
where $\chi_i^{v}$ is estimated from the transform matrices as described in Eq.~\ref{method:transform}.
After the above training, we obtain an optimized transform matrix for each viewpoint, which can effectively minimize the aligning errors introduced by the feed-forward mechanism.

%%%%%%%%%%%%%%%%%%%%%%%%% 4.3 Quantitative Results %%%%%%%%%%%%%%%%%%%%%%
%%%%%%%%%%%%%%%%%%%%%%%%%%%定量实验表格1
\begin{table}[t]
  \centering
  \resizebox{\linewidth}{!}{%
  \begin{tabular}{c c c c c} 
  \toprule
    Method & Scene-specific training & mIoU (\%) & mAcc (\%) & Total \\
    \midrule
    NVOS \cite{nvos}$\dagger$  & need      & 70.1 & 92.0 & - \\
    ISRF \cite{ISRF}         & need      & 83.8 & 96.4 & 840 s \\
    SGISRF \cite{SGISRF}$\dagger$ & need      & 84.5 & 97.2 & - \\
    SA3D \cite{sa3d}     & need      & 90.3 & 98.2 & \underline{780 s}  \\
    SAGA \cite{SAGA}         & need      & 90.9 & 98.3 & 2280 s \\
    OmniSeg3D \cite{ying2024omniseg3d}  & need      & 91.7 & 98.4 & 8220 s \\
    FlashSplat \cite{shen2024flashsplat}$\dagger$ & need      & \underline{91.8} &  \underline{98.6} & 1500 s \\
    WildSeg3D (ours) & \textbf{no need} & \textbf{94.1} & \textbf{99.0} & \textbf{30 s} \\
    \bottomrule
  \end{tabular}
  }
  \vspace{-2mm}
  \caption{\textbf{Quantitative results} on NVOS dataset. \textbf{Boldface} highlights the  best results and \underline{underline} the second-best. ``Total'' denotes the overall time from scene reconstruction to completing an interactive 3D segmentation. The symbol $\dagger$ denotes data taken from the respective references, as the code was not fully released.}
  \label{tab:1}
\end{table}
%
%%%%%%%%%%%%%%%%%%%%%%%%%%% 定量实验Figure  SPIn-NeRF 
\begin{table}[t]
  \centering
  \resizebox{\linewidth}{!}{%
  \setlength{\tabcolsep}{5pt}
  \begin{tabular}{c c c c c}
    \toprule
    Method & Scene-specific training & mIoU (\%) & mAcc (\%) & Time \\
    \midrule
    Single view\cite{sa3d} & need      & 74.6 & 95.5 & - \\
    MVSeg \cite{spinnerf}        & need      & 90.9 & 98.9 & 180-360 s \\
    ISRF \cite{ISRF}    & need      & 77.4 & 93.46 & 2-3 s \\
    SA3D \cite{sa3d}    & need      & 92.4 & \underline{98.9} & 120-600 s \\
    SAGA \cite{SAGA}    & need      & 88.0 & 98.5 & \underline{0.08-0.9 s} \\
    OmniSeg3D \cite{ying2024omniseg3d} & need  & \textbf{94.3} & \textbf{99.3} & 1-2 s \\
    WildSeg3D (ours) & \textbf{no need}   & \underline{94.0} & 98.6 & \textbf{0.005-0.02 s} \\
    \bottomrule
  \end{tabular}
  }
  \vspace{-2mm}
  \caption{\textbf{Quantitative results} on SPIn-NeRF dataset.}
  \label{tab:2}
\end{table}

\subsubsection{Multi-view Group Mapping}
WildSeg3D enables real-time 3D segmentation of target objects by using single-view prompts as input.
To efficiently map multi-view object masks into an aligned 3D space based on user prompts, we propose a multi-view group mapping method designed to optimize mask retrieval and integration.
The process begins by sorting all object masks within the mask cache for the current viewpoint in ascending order based on their area, with priority given to smaller, fine-grained objects.
Based on the user’s prompts, we then sequentially filter the relevant masks from the cache, appending the corresponding object IDs to the result set \( O \). 
Once the relevant masks are identified, we compute the union of these masks across all viewpoints in the dataset. Specifically, the unified mask for each viewpoint is defined as:
\begin{equation}
M=\left\{ M^v \mid M^v = \bigcup_{o \in O} m_o^v, v \in V \right\},
\label{method:USM}
\end{equation}
where $V$ represents the set of viewpoints, $m_o^v$ denotes the mask of object $o$ in viewpoint $v$, and $M$ is the collection of masks for each viewpoint after prompt-based retrieval.

We apply the transform matrices learned by DGA to convert the segmentation masks from all viewpoints in $M$ from pixel coordinates to aligned world coordinates. The resulting 3D segmentation is denoted as $\mathcal{P}$:
%%%%%% 公式here
\begin{equation}
\normalsize
\scalebox{0.742}{
    $\mathcal{P} =  \underset{v \in V}{\bigcup}  
    \left\{
    \mathbf{P}_v^{-1} \mathbf{K}_v^{-1} 
    \begin{pmatrix}
    x \mathbf{D}_v(x, y) \\
    y \mathbf{D}_v(x, y) \\
    \mathbf{D}_v(x, y)
    \end{pmatrix} 
    \;\middle|\;
    (x, y) \in M_v, M_v(x, y) = 1
    \right\}$,
}
\label{method:3dseg}
\end{equation}
where $\mathcal{P}$ represents a set of 3D point cloud coordinates for the object. Through this framework, \ourmodel~efficiently aggregates 2D masks from multiple viewpoints, enabling real-time interactive segmentation via the P2W (pixel to world coordinates) strategy.

%% file: sec/4_experiments.tex
\section{Experiments}

\subsection{Datasets}
To evaluate the effectiveness of our method, we conducted experiments on multiple benchmark datasets~\cite{nvos, spinnerf, knapitsch2017tanks, barron2022mip}.
The NVOS dataset provides a reference view with scribble annotations for the segmented targets, as well as a target view with the corresponding segmentation mask, both captured from frontal perspectives.
The SPIn-NeRF dataset, a 3D scene dataset, is annotated using the widely-adopted NeRF datasets~\cite{mildenhall2019local, mildenhall2021nerf, yen2022nerf, knapitsch2017tanks, fridovich2022plenoxels}. It is used to assess the performance of interactive 3D segmentation methods, including the evaluation of segmentation quality in more complex 3D environments.
For qualitative experiments, we used the NVOS\cite{nvos}, SPIn-NeRF~\cite{spinnerf}, T\&T\cite{knapitsch2017tanks}, and Mip-NeRF360~\cite{barron2022mip} datasets. These datasets were chosen to compare our method with existing approaches and to showcase the segmentation results produced by our method across different 3D scenes and viewpoints.
To evaluate segmentation accuracy and facilitate comparisons, we use mean Intersection over Union (mIoU) and mean Accuracy (mAcc) as primary metrics. Additionally, we assess both the training duration and the time required for interactive 3D segmentation on a single NVIDIA RTX 3090 GPU to evaluate the models' efficiency and real-time performance.
%

%%%%%%%%%%%%%%%%%%消融实验Table NVOS & SPIn-NeRF
\begin{table}[t]
  \centering
  \resizebox{\linewidth}{!}{%
  \begin{tabular}{c|c|c c|c c c@{}} % All columns centered
    \toprule
    Datasets & Scene & D-GA & D-DGA & M-GA & M-SGA & M-DGA \\
    \midrule
    \multirow{9}{*}{NVOS} 
    & fern & 82.5\% & \textbf{85.1\%} & \underline{94.1\%} & 71.7\% & \textbf{94.2\%} \\
    & flower & 90.3\% & \textbf{90.9\%} & 73.8\% & \underline{91.3\%} & \textbf{94.3\%} \\
    & fortress & 95.5\% & \textbf{96.3\%} & 95.5\% & \underline{96.0\%} & \textbf{96.8\%} \\
    & horns (center) & 88.4\% & \textbf{93.6\%} & \underline{93.0\%} & 92.9\% & \textbf{95.9\%} \\
    & horns (left) & 89.9\% & \textbf{95.2\%} & 94.8\% & \textbf{95.2\%} & \underline{95.0\%} \\
    & leaves & \textbf{65.8\%} & 61.9\% & 88.0\% & \underline{93.5\%} & \textbf{96.7\%} \\
    & orchids & 78.2\% & \textbf{82.6\%} & \underline{84.1\%} & 84.1\% & \textbf{93.5\%} \\
    & trex & 79.8\% & \textbf{80.2\%} & \underline{85.3\%} & 61.0\% & \textbf{86.4\%} \\
    \cline{2-7}
    & average & 83.8\% & \textbf{85.7\%} & \underline{88.6\%} & 85.7\% & \textbf{94.1\%} \\
    \midrule
    
    \multirow{10}{*}{SPIn-NeRF} 
    & fern & 82.5\% & \textbf{85.1\%} & \underline{94.1\%} & 71.7\% & \textbf{94.2\%} \\
    & fork & 85.4\% & \textbf{88.7\%} & \textbf{89.8\%} & 88.3\% & \underline{88.9\%} \\
    & fortress & 95.5\% & \textbf{96.3\%} & 95.5\% & \underline{96.0\%} & \textbf{96.8\%} \\
    & horns & 88.4\% & \textbf{93.6\%} & \underline{93.0\%} & 92.9\% & \textbf{95.9\%} \\
    & leaves & 65.8\% & 61.9\% & 88.0\% & \underline{93.5\%} & \textbf{96.7\%} \\
    & lego & 77.5\% & \textbf{79.4\%} & \underline{84.3\%} & 81.5\% & \textbf{92.8\%} \\
    & orchids & 78.2\% & \textbf{82.6\%} & \underline{84.1\%} & 84.1\% & \textbf{93.5\%} \\
    & pinecone & 84.2\% & \textbf{88.6\%} & 85.2\% & \underline{89.2\%} & \textbf{95.7\%} \\
    & room & \textbf{90.4\%} & 90.3\% & \underline{88.7\%} & 74.1\% & \textbf{91.5\%} \\
    & truck & 82.5\% & \textbf{82.6\%} & \underline{90.9\%} & 90.5\% & \textbf{93.7\%} \\
    \cline{2-7} % Draw line only for Scene and subsequent columns
    & average & 83.1\% & \textbf{84.9\%} & \underline{89.4\%} & 86.2\% & \textbf{94.0\%} \\
    \bottomrule
  \end{tabular}
  }
  \vspace{-2mm}
  \caption{\textbf{Effect of different alignment strategies} for DUSt3R-based and MASt3R-based 3D segmentation models on NVOS and SPIn-NeRF datasets.``D'' and ``M'' represent the pointmaps from DUSt3R and MASt3R, respectively. ``GA'' refers to global alignment, ``SGA'' denotes sparse global alignment in MASt3R, and ``DGA'' stands for our proposed dynamic global alignment.}
  \label{tab:ablation1}
\end{table}

\subsection{Quantitative Results}
\noindent\textbf{NVOS Dataset.}
To ensure experimental fairness, we adopted the evaluation approach used by models such as SAGA~\cite{SAGA}, utilizing the scribble annotations provided by the NVOS dataset~\cite{nvos} as input to generate 2D masks for the SAM model~\cite{sam}.
Additionally, we performed random point sampling on the scribble annotations of the reference view to acquire point prompts for segmentation. The results of our experiments on the NVOS dataset are presented in Table~\ref{tab:1}, where we compare the performance of our WildSeg3D framework against other state-of-the-art methods.
As shown in the table, WildSeg3D outperforms existing approaches in both mIoU and mAcc.
In terms of runtime efficiency, NeRF-based methods, including NVOS~\cite{nvos}, ISRF~\cite{ISRF}, SGISRF~\cite{SGISRF}, and SA3D~\cite{sa3d}, as well as 3DGS-based methods such as SAGA~\cite{SAGA}, OmniSeg3D~\cite{ying2024omniseg3d}, and FlashSplat~\cite{shen2024flashsplat}, typically require scene-specific training.
In contrast, our approach demonstrates strong generalization across diverse scenes, eliminating the need for scene-specific training. This advantage leads to significant efficiency improvements, with our model requiring only 3.8\% of the computation time of SA3D (NeRF-based) and 2\% of FlashSplat (3DGS-based), while maintaining high segmentation accuracy.

\noindent \textbf{SPIn-NeRF Dataset.}
The quantitative results of our experiments on the SPIn-NeRF dataset are presented in Table~\ref{tab:2}.
We evaluate the accuracy of our method by projecting the 3D segmentation masks onto the reference views and comparing them against the ground truth.
In the MVSeg method~\cite{hao2021edgeflow}, an interactive segmentation model is first employed to acquire object masks from a single view. These masks are then used to segment training views, which are treated as a video sequence and processed through a video instance segmentation model~\cite{caron2021emerging, wang2021survey} to generate 3D masks. These masks are further refined using a semantic NeRF model~\cite{mirzaei2022laterf,zhi2021place,zhi2021ilabel}.
For SA3D~\cite{sa3d}, interactive segmentation requires traversing all training views. Cross-view self-prompting is used to guide SAM in generating 2D masks, which are subsequently projected into 3D space using Mask Inverse Rendering. This process incurs significant time costs due to the need to process each view individually.
SAGA~\cite{SAGA} incorporates low-dimensional 3D features for each Gaussian, which are rendered into 2D feature maps through differentiable rasterization during training. This approach allows for contrastive training with the segmentation results generated by SAM.
In comparison to MVSeg~\cite{spinnerf} and SA3D~\cite{sa3d}, our method achieves comparable accuracy while requiring only a fraction—approximately one ten-thousandth—of the processing time. Moreover, when compared to SAGA, our approach not only demonstrates superior accuracy but also outperforms SAGA in terms of processing time. Overall, our method delivers exceptional performance, making real-time interactive 3D segmentation feasible in practical applications.

%%%%%%%%%%%%%%%%%%%%%%%%%% 定性实验figure-2
\begin{figure}[t]
  \centering
  \includegraphics[width=0.96\linewidth]{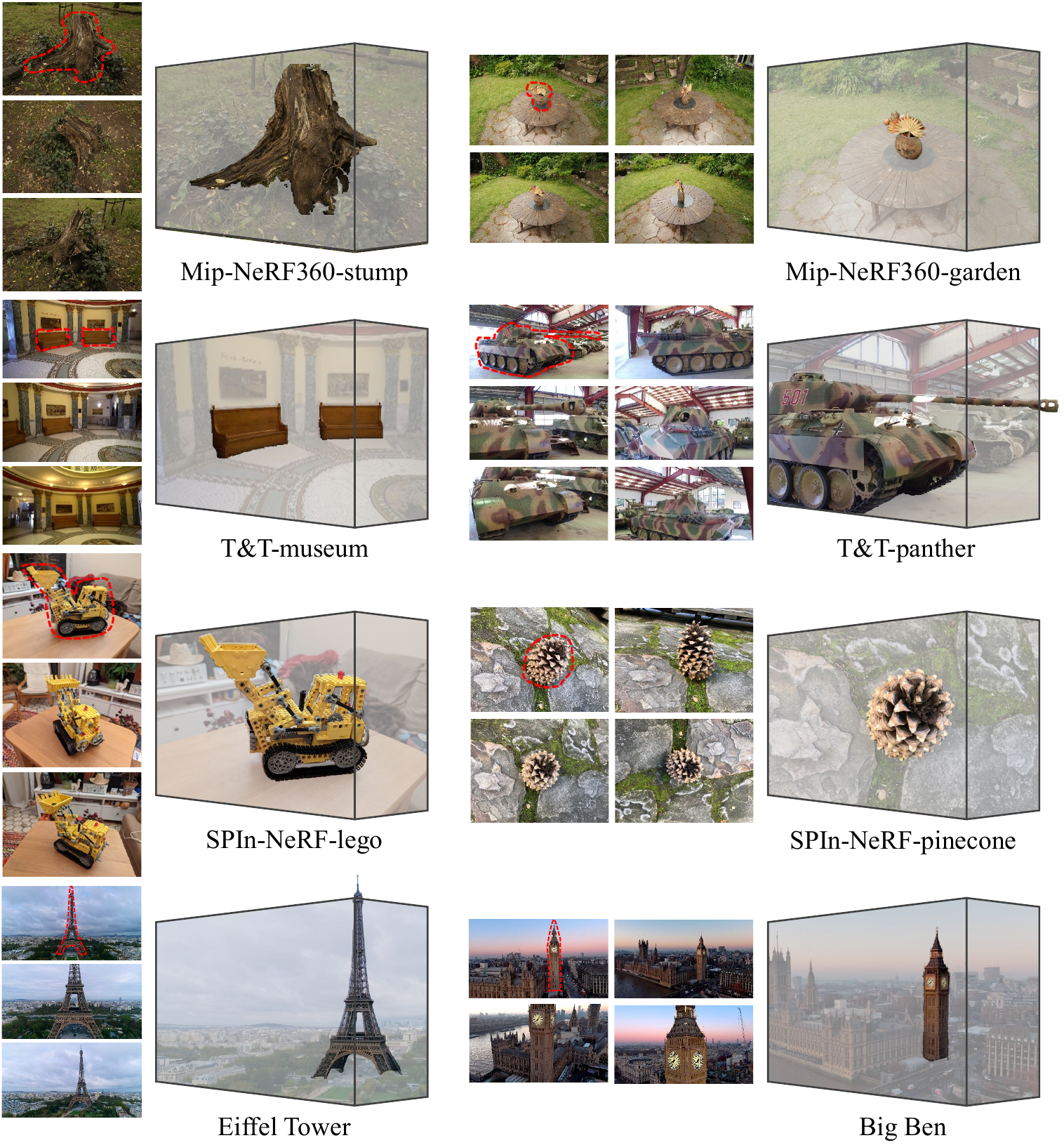}
  \vspace{-2.5mm}
  \caption{\textbf{Performance of~\ourmodel~on indoor and outdoor scenes.} For each scene, the left side shows the sparse views for reconstruction, with the segmentation target indicated by red dashed lines in the first view as prompts.}
\label{fig:qualitative2}
\end{figure}
%%%%%%%%%%%%%%%%%%%%%%%%% 4.4 Qualitative Results %%%%%%%%%%%%%%%%%%%%%%
\subsection{Qualitative Results}
We conduct a comparative analysis of our method against existing approaches and present qualitative experimental results. As shown in Figure~\ref{fig:qualitative1}, the first column illustrates the reconstruction time of our model for various scenes based on sparse views. The subsequent columns compare the segmentation results produced by our method with those from other approaches, along with the time taken to obtain 3D segmentation results after user input. The final column displays the segmentation results of ISRF~\cite{ISRF} across different scenes.
ISRF utilizes the TensoRF representation~\cite{chen2022tensorf} for scene training and rendering, incorporating DINO features~\cite{caron2021emerging} within each voxel to enable 2D-to-3D semantic matching. Segmentation is performed via nearest neighbor feature matching (NNFM). However, ISRF struggles with distinguishing semantically similar objects, as demonstrated in the Trex and Orchids scenes, where the method faces challenges in accurate segmentation.
SA3D~\cite{sa3d}, upon receiving user prompts, employs a cross-view self-prompting mechanism and SAM~\cite{sam} to generate 2D masks for each view. These masks are then mapped to 3D space via Mask Inverse Rendering, with the process repeated for each user interaction. This approach incurs high computational costs due to the need for repeated processing.
SAGA~\cite{SAGA} and OmniSeg3D~\cite{ying2024omniseg3d} transform 2D masks into 3D features and require additional training for each 3D Gaussian. SAGA relies on a collection of loss functions during training, whereas OmniSeg3D uses hierarchical contrastive learning to map discontinuous multi-view segmentations to consistent 3D features.
In contrast, our method achieves superior performance with high-quality transform matrices learned through DGA, which map 2D views into an aligned 3D coordinate system. The MGM module then enables fast retrieval of 2D masks from multiple views, which are subsequently transformed into the unified 3D space. This design results in significantly improved computational efficiency, allowing our method to achieve faster interactive segmentation compared to existing approaches.

As demonstrated in Figure~\ref{fig:qualitative2}, our method does not require scene-specific training, enabling near real-time 3D interactive segmentation for arbitrary scenes. We showcase this capability using indoor and outdoor scenes from the Mip-NeRF360~\cite{barron2022mip}, T\&T~\cite{knapitsch2017tanks}, SPIn-NeRF~\cite{spinnerf} datasets, and two additional scenes in the wild.
By leveraging only sparse views, our method is able to perform both scene reconstruction and near real-time 3D segmentation, underscoring its versatility and ability to segment objects in diverse, uncontrolled environments.

\subsection{Ablation Studies}
\label{sec:ablation}
\textbf{Comparison of Different Viewpoint Counts.}
%我们对输入到WildSeg3D的视角数量进行了消融实验，我们固定了全局对齐的训练步数，因此视角数量太多mIoU有所下降。考虑综合性能，前面实验我的随机选取5个视角。在SPIn-NeRF数据集上，当视角数量为7时，我们的精度超越了Table 2(5个视角)中的OmniSeg3D。
Table~\ref{tab:4} presents an ablation study on the number of viewpoints input into WildSeg3D. Considering overall performance, we randomly select five viewpoints in prior experiments. On the SPIn-NeRF dataset, our method outperforms OmniSeg3D when seven viewpoints are selected, despite the results in Table~\ref{tab:2}.

%DGA利用经过dynamic adjustment function权重，进行梯度下降，to conduct alignment across multi views3.
\noindent \textbf{Effect of Dynamic Global Alignment.}
%DGA employs dynamically adjusted weights through an dynamic adjustment function for gradient descent optimization, aligning multiple views, which substantially reduces 3D alignment errors. We conduct ablation experiments on the NVOS and SPIn-NeRF datasets. 
% To further validate the effectiveness of DGA, we also integrated it into DUSt3R, where the matching maps in DGA are replaced with SIFT~\cite{sift}. As shown in Table~\ref{tab:ablation1} ,whether for models based on DUSt3R or MASt3R, incorporating DGA significantly enhances performance compared to traditional GA training approaches. 
To further validate the effectiveness of DGA, we also integrate it into DUSt3R, replacing the matching maps with SIFT~\cite{sift}. As shown in Table~\ref{tab:ablation1}, incorporating DGA significantly improves performance in both DUSt3R and MASt3R models compared to traditional GA training methods.

\begin{figure}[t]
  \centering
  \includegraphics[width=0.98\linewidth]{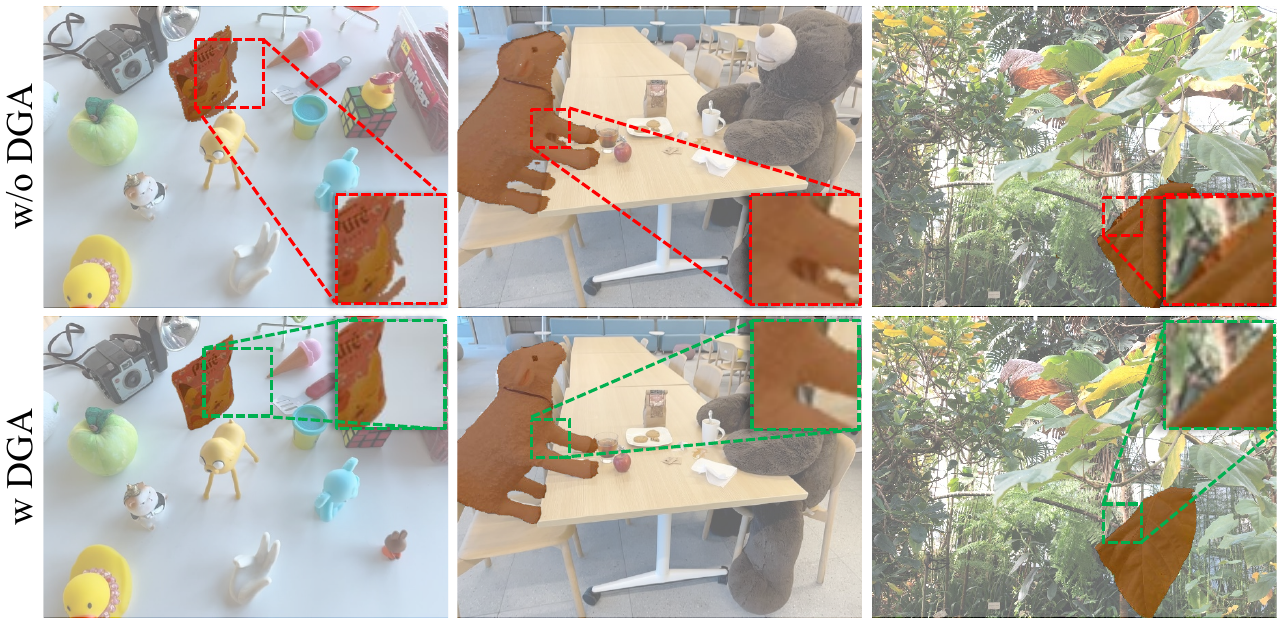}
  \vspace{-3mm}
  \caption{\textbf{Visualization} of ablation experiments on DGA.}
  % 关于DGA模块的可视化消融实验
\label{fig:abllation_DGA}
\end{figure}

Furthermore, as shown in Figure~\ref{fig:abllation_DGA}, comparative visualizations on the NVOS and LERF~\cite{LERF} datasets demonstrate that DGA effectively reduces 3D alignment errors, minimizing blurred details and background confusion.
%Furthermore, 如Figure xxx所示，我们在NVOS和LEFR数据集上的可视化对比说明了DGA可以有效减轻3D alignment error 导致的blurred detail and potential background confusion

%%%%%%%%%%%%%%%%%%%%%%%%% 消融实验 多视角图片数量
%%%%%%%%%%%%%%%%%%%%%%%%%%% 定量实验Figure  SPIn-NeRF 
\begin{table}[t]
  \centering
  \resizebox{\linewidth}{!}{%
  \setlength{\tabcolsep}{5pt}
  \begin{tabular}{c c c c c c} 
    \toprule
    Number of Views & 3 & 4 & 5 & 7 & 9 \\
    \midrule
    NVOS & 89.8\% & 92.9\% & \textbf{94.1\%} & 94.0\% & 93.8\% \\
    SPIn-NeRF & 87.8\% & 92.2\% & 94.0\% & \textbf{94.9\%} & 94.7\% \\
    Total Time  & 15 s    & 18 s    & 23 s    & 35 s    & 50 s \\
    \bottomrule
  \end{tabular}
  }
  \vspace{-2mm}
  \caption{\textbf{Ablation Experiments} on the Impact of Viewpoint Quantity on mIoU.}
  \label{tab:4}
\end{table}

%%%%%%%%%%%%%%%%%%%%%%%%%% 消融实验 w/o DGA

\noindent \textbf{Different Confidence adjustment functions.}
%As shown in Figure~\ref{fig:functions},
Figure~\ref{fig:functions} compares the confidence adjustment function in DGA with other adjustment functions, defined as: $A_{\theta}(x) = \frac{\arctan(\theta \cdot (x - 0.5))}{\pi} + 0.5$, $S_{\delta}(x) = \frac{1}{1 + e^{-\delta \cdot (x - 0.5)}}$, $L_{\beta}(x) = \frac{\log(1 + \beta \cdot x)}{\log(1 + \beta)}$,
where \( \theta \), \( \delta \), and \( \beta \) control the scaling of the Arctan, Sigmoid, and Logarithmic functions, respectively.

% The experimental results are presented in Figure~\ref{fig:functions}, where appropriate hyperparameters were set for each function. Remarkably, our confidence adjustment function achieved the best performance.
Our function performs adaptive adjustments centered at 0.5 confidence, improving the weight of hard-to-match points while ensuring smooth transitions across both low and high confidence levels, achieving the best performance. 
% \setlength{\tabcolsep}{16pt} % 调整列间距
% \begin{table}[t]
%   \centering
%   \setlength{\tabcolsep}{15pt}
%   \resizebox{0.35\textwidth}{0.05\textheight}{%
%   \begin{tabular}{ccc} % Center all columns
%     \toprule
%     Functions & mIoU & mAcc \\
%     \midrule
%     Arctan & 91.3\% & 97.8\% \\
%     Sigmoid & 91.3\% & 97.8\% \\
%     Logarithmic & 89.5\% & 97.4\% \\
%     Ours & \textbf{94.1\%} & \textbf{99.0\%} \\
%     \bottomrule
%   \end{tabular}
%   }
%   \vspace{-2mm}
%   \caption{Ablation Experiments on the NVOS Dataset Applying Different Functions to DGA.}
%   \label{tab:ablation2}
% \end{table}

%% file: sec/5_conclusion.tex
\section{Conclusion}
In this paper, we introduce WildSeg3D, a feed-forward method that enables real-time interactive 3D segmentation from 2D images without scene-specific pre-training.
With Dynamic Global Aligning, WildSeg3D learns high-quality transform matrix for each view, aligning 2D images to an aligned coordinate system with minimal alignment error.
Additionally, we propose a mask cache, which stores masks from multiple views consistently. 
For real-time interactive segmentation, MGM uses a search strategy to retrieve target masks and map them into an aligned 3D space, responding promptly to user inputs.
Through extensive experiments, WildSeg3D demonstrates a significant speedup over existing methods while maintaining high accuracy, highlighting its potential for downstream tasks.
\begin{figure}[t]
  \centering
  \includegraphics[width=0.98\linewidth]{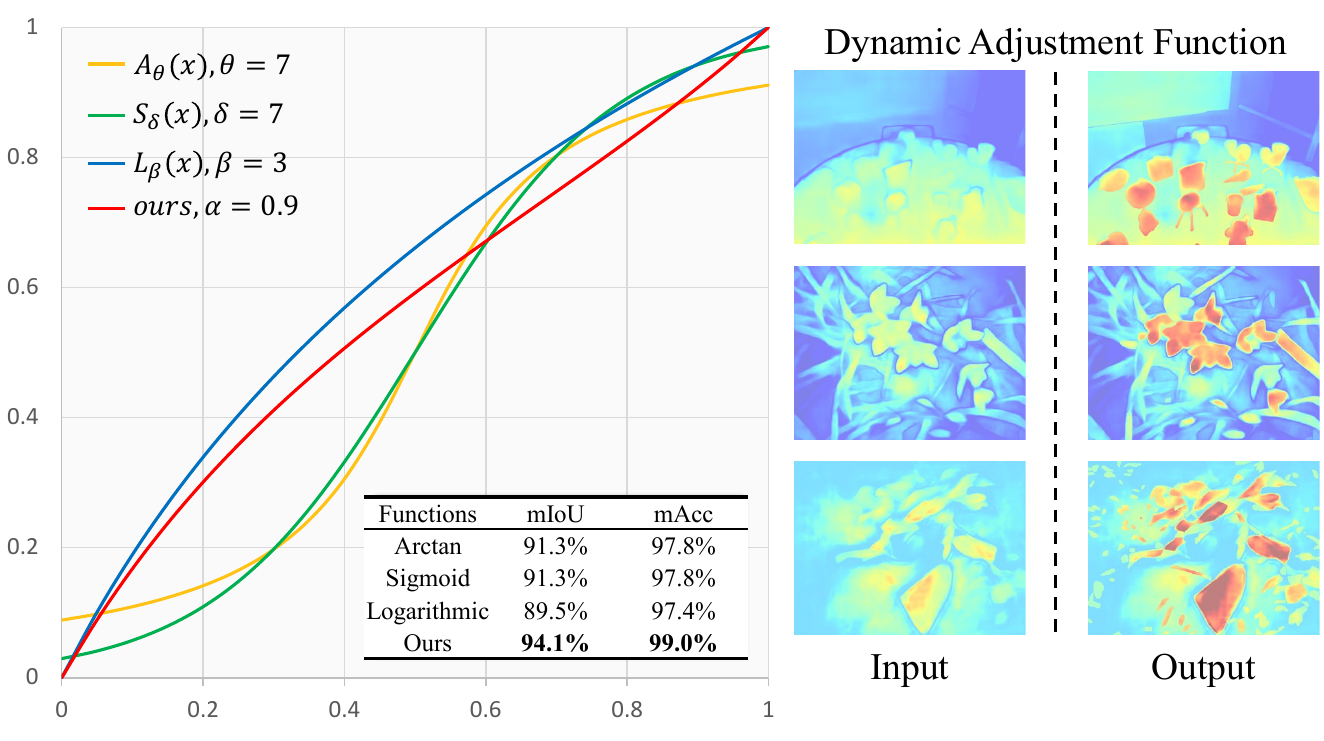} % 调整这里的宽度
  \vspace{-2mm}
  \caption{\textbf{Ablation Experiments} on NVOS Dataset. Left: visualization of different confidence adjustment functions for DGA. Right: Results of these functions.}
  \label{fig:functions}
\end{figure}

%% file: sec/X_suppl.tex
\clearpage
\setcounter{page}{1}
\maketitlesupplementary

\section{IMPLEMENTATIONAL DETAILS}

% 我们的工作是基于预训练好的MASt3R。在DGA模块中,我们的dynamic adjustment function包含两个超参数，在定量与定性实验中，我们分别将两个超参数设置为0.5和0.9。我们的方法基于sparse view构建3D场景, 针对dynamic aligning loss的训练环节，我们设置了500个循环。对于一个场景，从预处理步骤到全局优化对齐的场景重建再到基于用户提示的3D分割,整个过程运行在一张NVIDIA RTX 3090 GPU上，仅需大约30s。
In the DGA module, the dynamic adjustment function incorporates two hyperparameters $\alpha_p$ and $\alpha_n$, which are set to 0.5 and 0.9 in both quantitative and qualitative experiments. Our approach constructs 3D scenes using sparse views, and for the training phase of the dynamic aligning loss, we employ 500 iterations. For each scene, the entire workflow, from pre-processing to global alignment reconstruction and user-prompted 3D segmentation, runs on a single NVIDIA RTX 3090 GPU in approximately 30 seconds.

\section{EXPERIMENTAL DETAILS}

\subsection{More Qualitative Evaluation}
\begin{figure*}[t]
  \centering
  \includegraphics[width=1\linewidth]{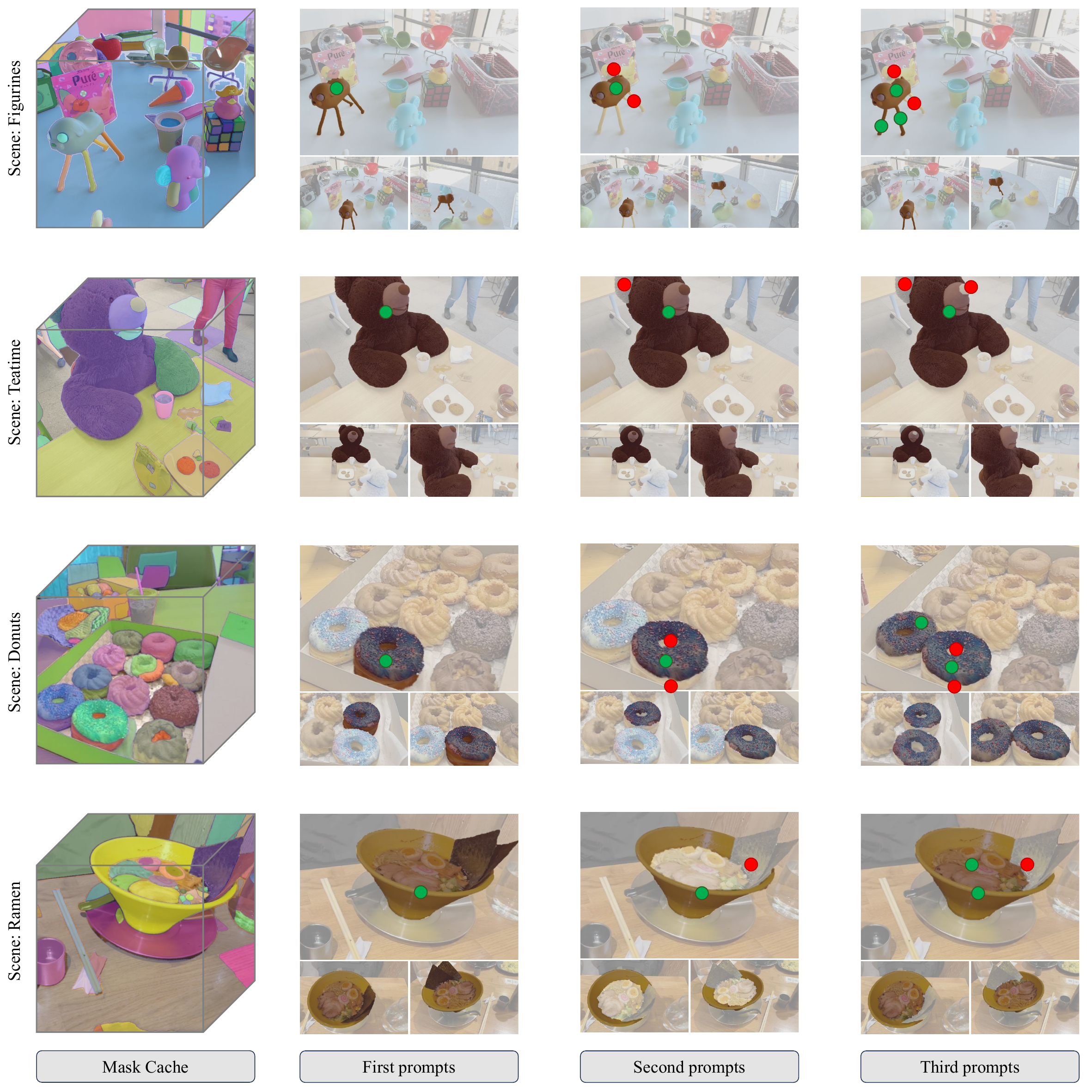}
  \vspace{-2.5mm}
  \caption{\textbf{Visualization of mask cache and interactive segmentation on the LERF dataset.} In each row, the segmentation masks in the first column represent the stored masks in our mask cache. Each scene undergoes three interactive refinement steps, with three distinct viewpoints displayed per prompt. Green and red points denote user-provided positive and negative prompts, respectively.}
\label{fig:interactive}
\end{figure*}

% 我们基于Mask Cache实现了实时交互式分割，支持用户交互进行multi-round segmentation。进一步地，我们基于SAM2展示了用户可以add further interaction to refine and make the segmentation mask better, 如图1所示。
We implemented real-time interactive segmentation based on mask cache, supporting user interaction for multi-round segmentation. Furthermore, leveraging SAM2, we demonstrate that users can add further interactions to refine the segmentation mask. Visualization results based on the LERF~\cite{LERF} dataset are shown in Figure~\ref{fig:interactive}.

\begin{figure*}[t]
  \centering
  \includegraphics[width=1\linewidth]{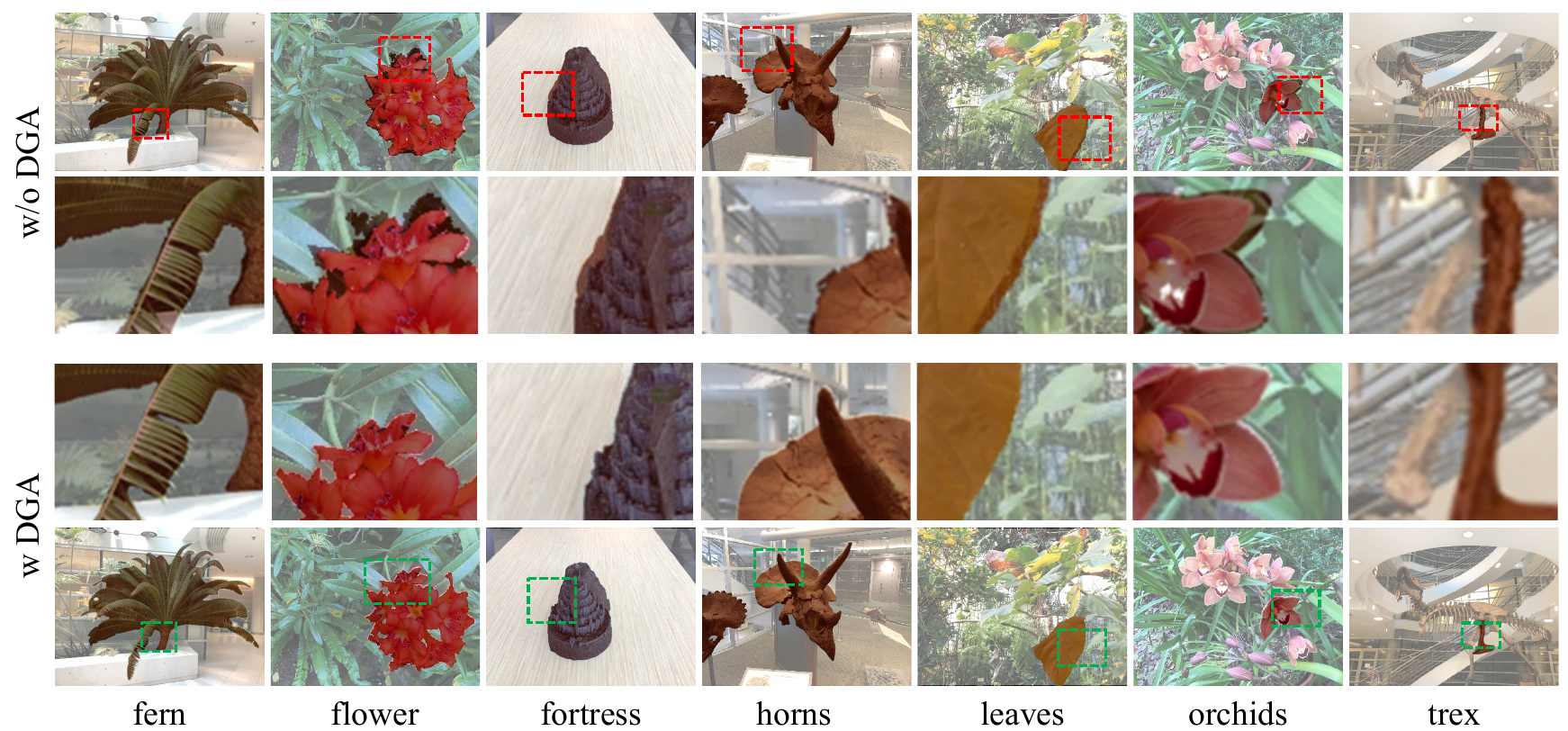}
  \vspace{-2.5mm}
  \caption{\textbf{Visualization of ablation experiment on the NVOS dataset.} In each column, the images depicted in the top and bottom rows illustrate the segmentation results without and with DGA, respectively. The second and third rows highlight zoomed-in views of the areas within the red and green dashed boxes.}
\label{fig:A1}
\end{figure*}

\begin{figure*}[ht]
  \centering
  \includegraphics[width=1\linewidth]{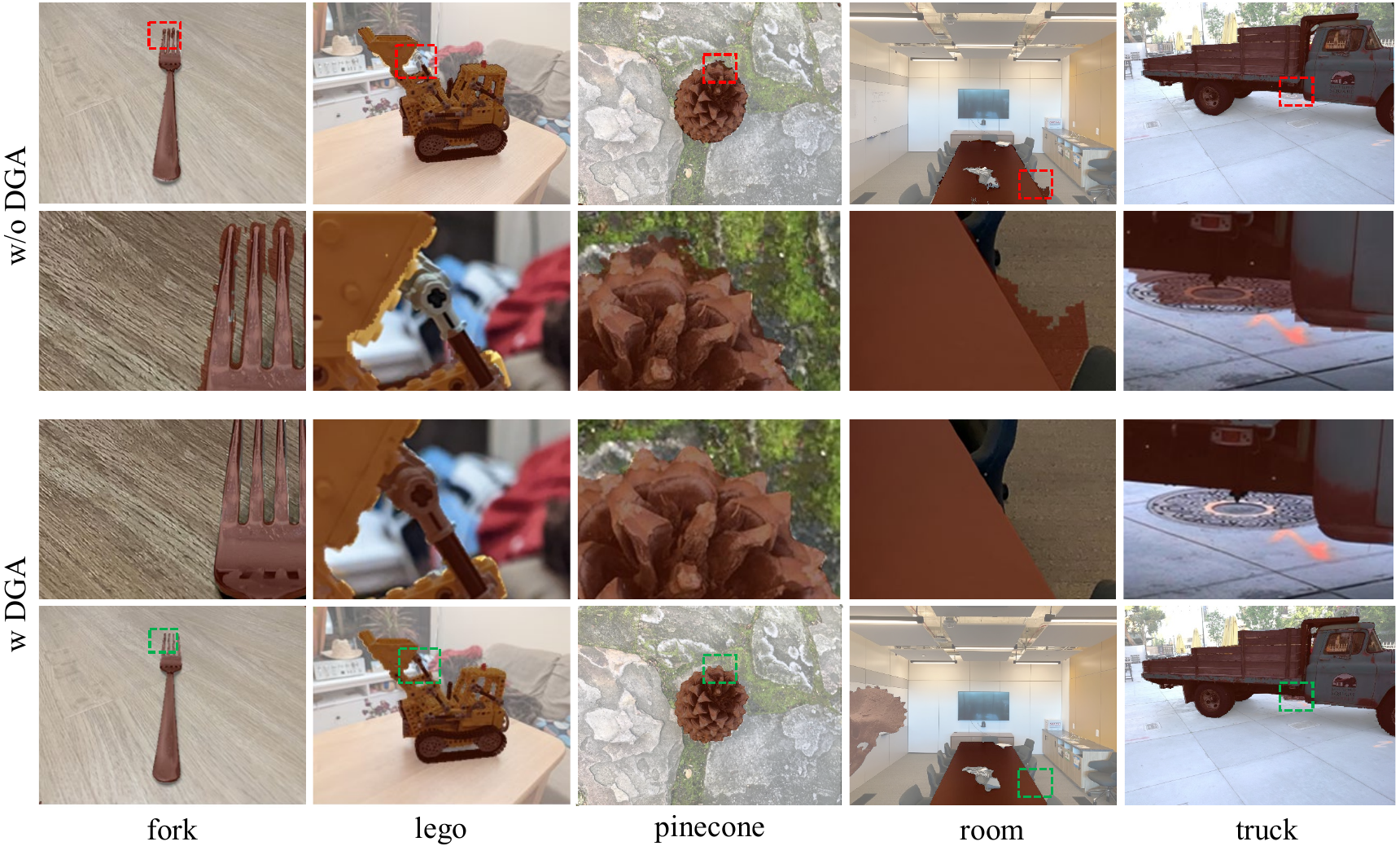}
  \vspace{-2.5mm}
  \caption{\textbf{Visualization of ablation experiment on the SPIn-NeRF
 dataset.}}
\label{fig:A2}
\end{figure*}

\begin{figure*}[ht]
  \centering
  \includegraphics[width=1\linewidth]{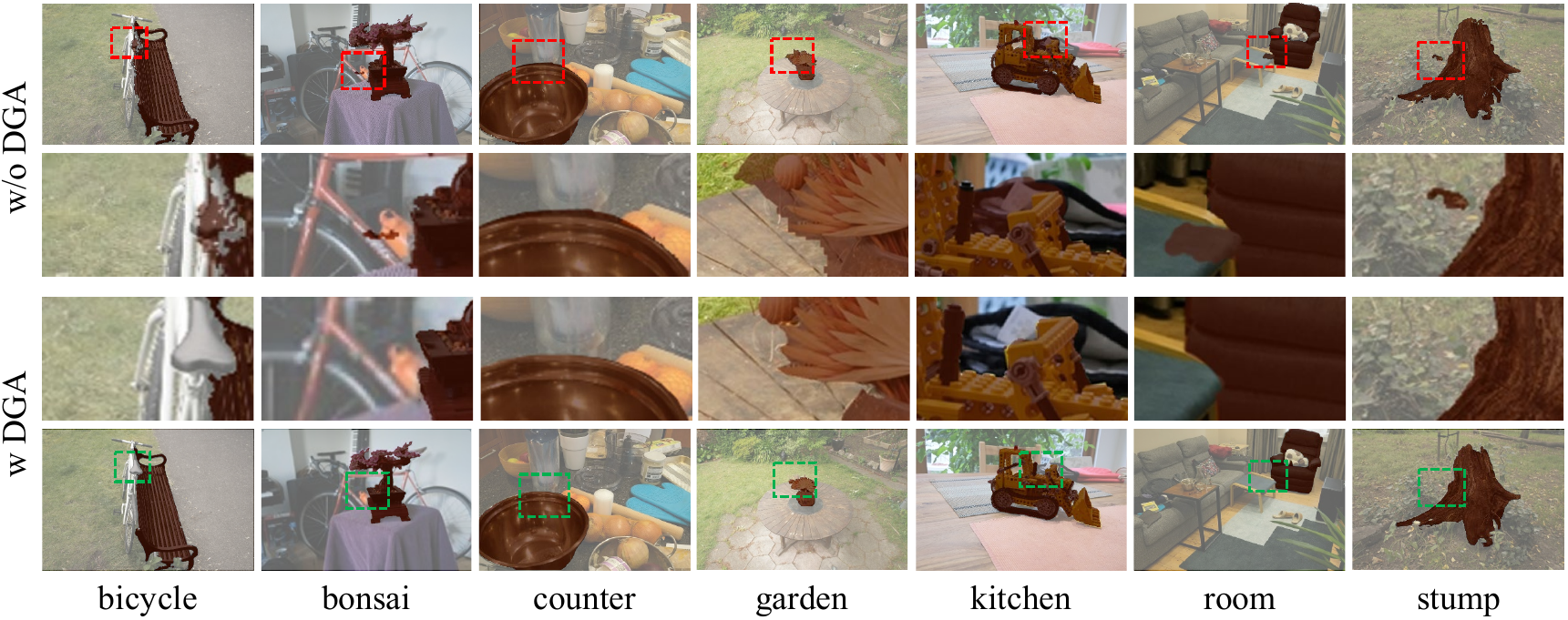}
  \vspace{-2.5mm}
  \caption{\textbf{Visualization of ablation experiment on the Mip-NeRF360 dataset.}}
\label{fig:A3}
\end{figure*}

\begin{figure*}[ht]
  \centering
  \includegraphics[width=1\linewidth]{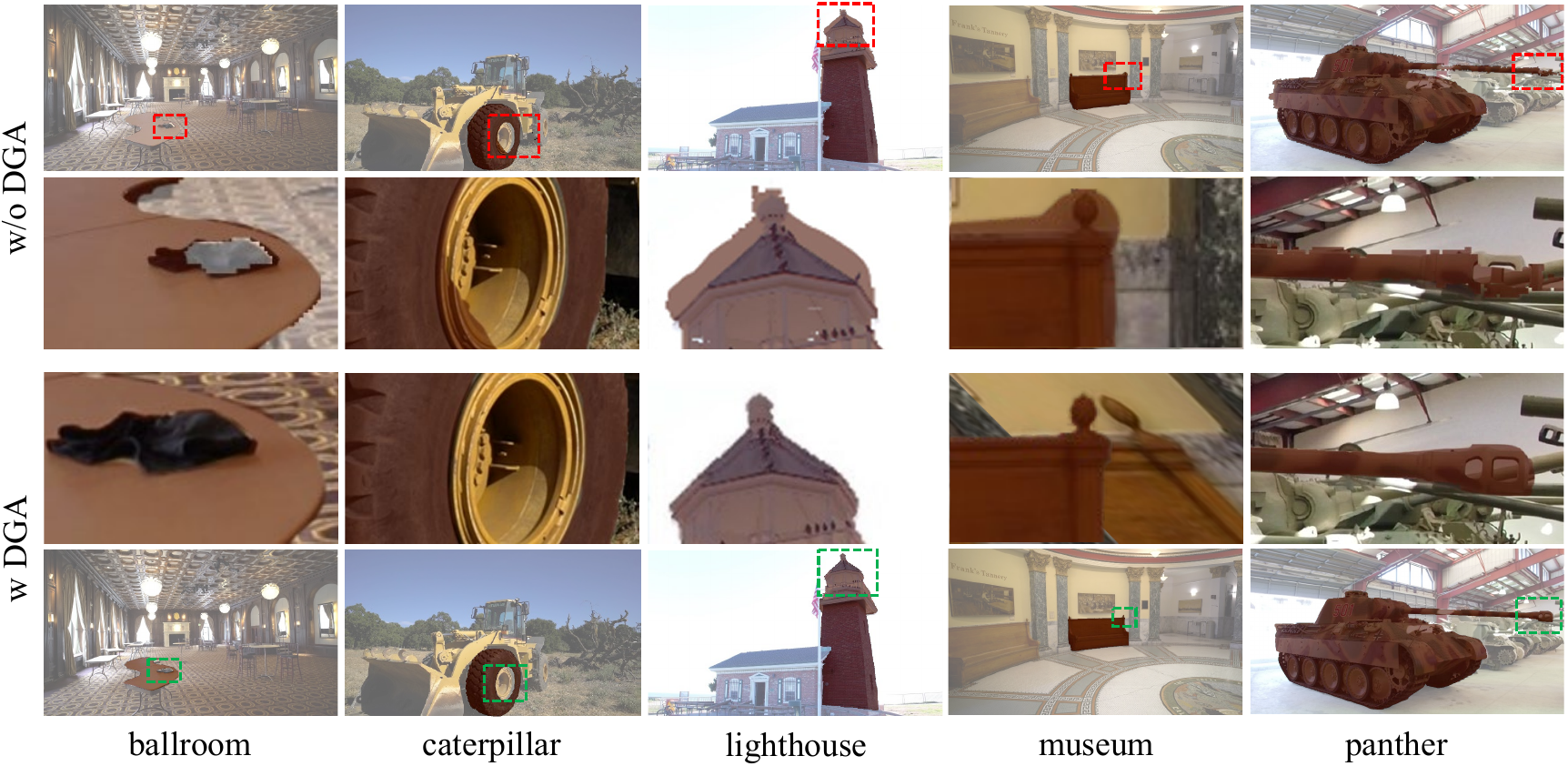}
  \vspace{-2.5mm}
  \caption{\textbf{Visualization of ablation experiment on the T\&T
 dataset.}}
\label{fig:A4}
\end{figure*}

To validate the effectiveness of our WildSeg3D, we conducted visualization-based ablation studies on segmentation tasks. 
As outlined in Sec.~\ref{sec:ablation}, DGA demonstrates its ability to effectively reduce 3D alignment errors by mitigating the influence of misaligned points and redundant background information. 
This improvement significantly enhances the clarity of object boundaries while minimizing artifacts such as blurred details and background confusion.
The efficiency of DGA has previously been demonstrated. Furthermore, Figures ~\ref{fig:A1},~\ref{fig:A2},~\ref{fig:A3}, and~\ref{fig:A4} present visualization experiments conducted on the NVOS~\cite{nvos}, SPIn-NeRF~\cite{spinnerf}, Mip-NeRF360~\cite{barron2022mip}, and T\&T~\cite{knapitsch2017tanks} datasets, respectively. 
%没有DGA，在全局对齐步骤中，将来自多视角的pointmaps对齐到统一坐标系时容易受到 redundant background points的影响并且对3D points across different views, which have different difficulty of matching施以相同的对齐权重，会累积3D alignment errors。

Without DGA, aligning pointmaps from multiple views to a unified coordinate system is significantly affected by the presence of redundant background points, which adversely affect the global aligning accuracy. This issue becomes particularly pronounced in complex scenes.
Moreover, directly using confidence scores as aligning weights for 3D points across different views, which vary in matching difficulty, can lead to the accumulation of 3D alignment errors.
Notably, the incorporation of DGA enables our method to accurately demarcate object boundaries while suppressing the influence of background pixels and dynamically adjusting aligning weights.

%%%%%%%%%%%%%%%%%%%%%%%%%%% 消融实验 alpha_p
\begin{table}[t]
  \centering
  \resizebox{\linewidth}{!}{%
  \setlength{\tabcolsep}{10pt}
  \begin{tabular}{c|ccccc}
    \toprule
    \multirow{2}{*}{Scene} & \multicolumn{5}{c}{$\alpha_p$} \\
           & -0.9 & -0.5 & 0 & 0.5 & 0.9 \\ \midrule
    fern & 93.9\% & 93.9\% & 94.2\% & 94.2\% & 94.1\% \\
    flower & 91.0\% & 91.1\% & 91.2\% & 91.1\% & 91.0\% \\
    fortress & 97.8\% & 97.8\% & 97.7\% & 97.7\% & 97.6\% \\
    horns (left) & 95.2\% & 95.1\% & 95.2\% & 95.2\% & 95.1\% \\
    horns (center) & 94.1\% & 93.9\% & 93.5\% & 93.5\% & 93.2\% \\
    leaves & 90.9\% & 91.3\% & 92.2\% & 94.0\% & 94.1\% \\
    orchids & 89.4\% & 90.1\% & 90.0\% & 91.0\% & 90.6\% \\
    trex & 39.7\% & 79.9\% & 84.5\% & 86.4\% & 86.3\% \\
    \midrule
    average & 86.5\% & 91.6\% & 92.3\% & \textbf{92.9\%} & 92.8\% \\
    \bottomrule
  \end{tabular}
  }
  \vspace{-2mm}
  \caption{Ablation experiment on the NVOS dataset for the adjustment factor $\alpha_p$.}
  \label{tab:alpha_p}
\end{table}

%%%%%%%%%%%%%%%%%%%%%%%%%%% 消融实验 alpha_n
\begin{table}[t]
  \centering
  \resizebox{\linewidth}{!}{%
  \setlength{\tabcolsep}{10pt}
  \begin{tabular}{c|ccccc}
    \toprule
    \multirow{2}{*}{Scene} & \multicolumn{5}{c}{$\alpha_n$} \\
           & -0.9 & -0.5 & 0 & 0.5 & 0.9 \\ \midrule
    fern & 73.2\% & 75.3\% & 94.2\% & 94.0\% & 94.2\% \\
    flower & 84.0\% & 40.4\% & 91.2\% & 89.4\% & 89.5\% \\
    fortress & 98.2\% & 78.6\% & 97.7\% & 97.6\% & 97.7\% \\
    horns (left) & 96.3\% & 96.4\% & 95.2\% & 95.1\% & 95.0\% \\
    horns (center) & 35.5\% & 32.2\% & 93.5\% & 92.9\% & 93.0\% \\
    leaves & 93.8\% & 66.8\% & 92.2\% & 96.3\% & 96.6\% \\
    orchids & 92.3\% & 91.2\% & 90.0\% & 90.1\% & 90.3\% \\
    trex & 82.0\% & 78.6\% & 84.5\% & 86.3\% & 86.4\% \\
    \midrule
    average & 81.9\% & 69.9\% & 92.3\% & 92.7\% & \textbf{92.8\%} \\
    \bottomrule
  \end{tabular}
  }
  \vspace{-2mm}
  \caption{Ablation experiment on the NVOS dataset for the adjustment factor $\alpha_n$.}
  \label{tab:alpha_n}
\end{table}

%%%%%%%%%%%%%%%%%%%%%%%%%%%交互式3D分割 可视化
% 在每一行，第一列的分割掩码表示我们的mask cache中存放的分割掩码. 在每一个场景下,我们分别展示了三个步骤的交互式来修改分割mask. 绿色点和红色点分别表示用户输入的positive和negative prompts.

\subsection{Evaluation on Adjustment Factor $\alpha$}
Tables~\ref{tab:alpha_p} and~\ref{tab:alpha_n} present the results of our ablation experiments conducted on the NVOS dataset. For each scene, we select five views as training views and use the mask segmented by SAM2 from one reference view as the ground truth for evaluation.  
With the Adjustment Factor $\alpha$ ranging from $[-1, 1]$, we project the 3D segmentation results from all scenes onto the  reference views and compute the mIoU with their ground truth masks.
The Adjustment Factor involves two hyperparameters, $\alpha_p$ and $\alpha_n$, representing matching and non-matching points, respectively.
To evaluate the impact of each hyperparameter on segmentation performance, we systematically investigate their individual contributions. For example, to analyze the influence of $\alpha_p$, we fix $\alpha_n$ at 0, and vice versa.  
Among all values of $\alpha_p$ and $\alpha_n$, the settings of $\alpha_p = 0.5$ and $\alpha_n = 0.9$ achieve the optimal mean IoU of 92.9\% and 92.8\%, respectively. 
\section{Segmentation in the Wild}
\begin{figure*}[t]
  \centering
  \includegraphics[width=1\linewidth]{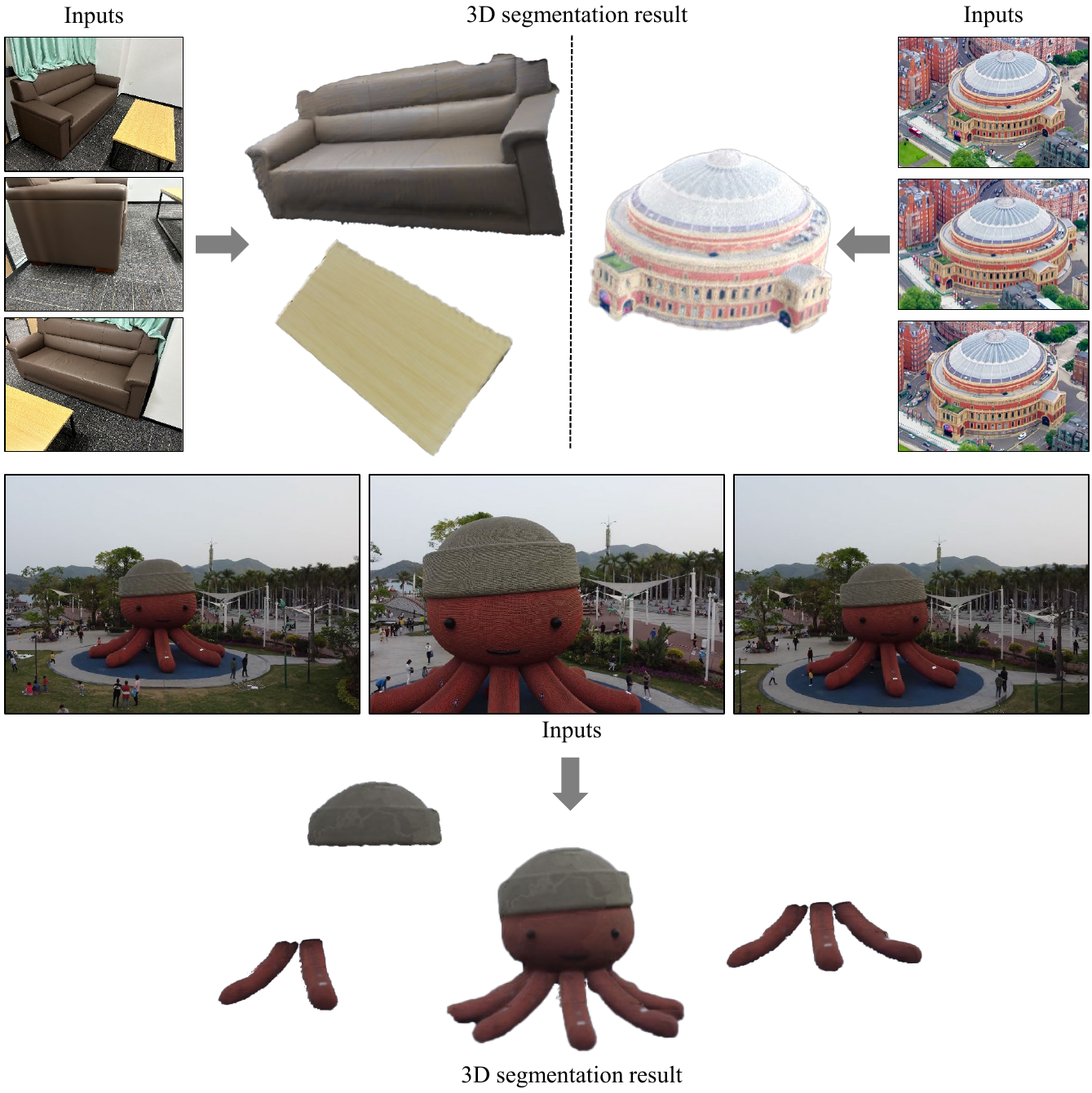}
  \vspace{-2.5mm}
  \caption{Visualization of WildSeg3D’s segmentation results on scenes in the wild with highly sparse views.}
\label{fig:wild}
\end{figure*}

Our method demonstrates the capability of real-time segmentation in arbitrary scenes. As shown in Figure~\ref{fig:qualitative2}, we perform segmentation on both indoor and outdoor scenes. Notably, the "Eiffel Tower" and "Big Ben" scenes are sourced from outdoor aerial videos, where a selection of frames is extracted and used as images for 3D segmentation. To showcase WildSeg3D's capability in segmenting arbitrary objects, we evaluate it on additional scenes, as shown in Figure~\ref{fig:wild}. Our method achieves robust performance on diverse real-world scenes with highly sparse views, completing reconstruction and interactive segmentation within 10 seconds in unconstrained environments.
%我们的方法具有在任意场景下的实时分割能力，如Figure.3所示,我们对室内外的场景进行了分割,其中的"Eiffel Tower"和"Big Ben"两个场景来自于室外的航拍视频，我们分别从视频中截取了一些帧用于3D分割。为了进一步展示WildSeg3D的in the wild的对任意目标的分割能力,我们选取了一部分场景，如图xxx所示。

% 在NVOS数据集上针对调节因子$\alpha_p$的消融实验

% \label{sec:rationale}
% % 
% Having the supplementary compiled together with the main paper means that:
% % 
% \begin{itemize}
% \item The supplementary can back-reference sections of the main paper, for example, we can refer to \cref{sec:intro};
% \item The main paper can forward reference sub-sections within the supplementary explicitly (e.g. referring to a particular experiment); 
% \item When submitted to arXiv, the supplementary will already included at the end of the paper.
% \end{itemize}
% % 
% To split the supplementary pages from the main paper, you can use \href{https://support.apple.com/en-ca/guide/preview/prvw11793/mac#:~:text=Delete%20a%20page%20from%20a,or%20choose%20Edit%20%3E%20Delete).}{Preview (on macOS)}, \href{https://www.adobe.com/acrobat/how-to/delete-pages-from-pdf.html#:~:text=Choose%20%E2%80%9CTools%E2%80%9D%20%3E%20%E2%80%9COrganize,or%20pages%20from%20the%20file.}{Adobe Acrobat} (on all OSs), as well as \href{https://superuser.com/questions/517986/is-it-possible-to-delete-some-pages-of-a-pdf-document}{command line tools}.